\documentclass[letterpaper, 10 pt, conference]{ieeeconf}  %

\IEEEoverridecommandlockouts                              %

\let\labelindent\relax
\usepackage{enumitem}
\usepackage{multirow}
\usepackage{amsmath}
\usepackage{amssymb}
\usepackage{caption}
\usepackage{dsfont}
\usepackage{graphicx}
\usepackage{booktabs}
\usepackage{siunitx}
\usepackage{wrapfig}
\usepackage{balance}
\usepackage[final]{microtype}
\usepackage{xcolor}
\title{\LARGE \bf
On the Role of the Action Space in Robot Manipulation\\ Learning and Sim-to-Real Transfer
}
\newcommand\hl[1]{\textcolor{black}{#1}}
\newcommand\hll[1]{\textcolor{black}{#1}}

\author{Elie Aljalbout$^{*}$, Felix Frank$^{*}$, Maximilian Karl, and Patrick van der Smagt$^{1}$%
\thanks{$^{1}$All authors are with the Volkswagen Group, Machine Learning Research Lab, Munich, Germany. $^{*}$Shared first authorship.}
}
\begin{document}

\maketitle
\thispagestyle{empty}
\pagestyle{empty}

\begin{abstract}

We study the choice of action space in robot manipulation learning and sim-to-real transfer.
We define metrics that assess the performance, and examine the emerging properties in the different action spaces.
We train over 250 reinforcement learning~(RL) agents in simulated reaching and pushing tasks, using 13 different control spaces.
\hl{The choice of spaces spans combinations of common action space design characteristics.}
We evaluate the training performance in simulation and the transfer to a real-world environment.
We identify good and bad characteristics of robotic action spaces and make recommendations for future designs.
Our findings have important implications for the design of RL algorithms for robot manipulation tasks, and highlight the need for careful consideration of action spaces when training and transferring RL agents for real-world robotics.
\end{abstract}

\section{Introduction}
Robot reinforcement learning (RL) provides a way to acquire manipulation skills without requiring explicit programming of task plans or models of task objects.
This property can enable robot manipulation to be feasible in a wide variety of tasks and applications.
However, due to the complexity of real-world manipulation tasks, making RL work in these domains can be very challenging.
During policy search, the robot acts almost randomly for a large period of time in order for the agent to explore its state and action spaces.
This can be a very dangerous and expensive process and should be approached with care.
Alternatively, such skills can be safely learned in simulation and transferred later to the real world\hl{~\cite{zhao2020sim, muratore2022robot}}.
In recent years, many approaches have been proposed to overcome the sim-to-real gap~\cite{Heiden2020AugmentingDS,hwangbo2019learning,Borrego2018MindTG}.
These methods have been shown to enable the successful transfer of manipulation policies learned in simulation.
Despite all efforts, the gap could not yet be fully bridged or even understood.
This lack of understanding hinders the applicability of sim-to-real transfer to a wide range of applications.

The majority of previous studies focused on the state \hl{and observation spaces} and perception aspects of robot policy learning and sim-to-real transfer.
\textsl{Our} goal is to understand the \textit{action} space and \textit{control} aspects of this problem.

To provide more context, in recent years there has been a shift toward embedding well-established control principles into the action space, such as proportional-derivative-controlled joint positions~\cite{Rudin2021LearningTW} or impedance control in end-effector space \cite{martin2019variable}.
This was in contrast to earlier interests in end-to-end policies, which directly output the lowest level possible of control commands such as joint torques~\cite{wahlstrom2015pixels}.
In this recent trend in robot learning, the policy outputs a higher-level control command, such as desired joint velocities, which are then fed to a low-level hand-engineered controller that handles the low-level control.
Such an approach reduces the complexity of the policy, effectively making learning simpler and more efficient.
Nevertheless, the introduction of these low-level controllers can fundamentally change the nature of the problem, in some cases even violating some basic assumptions made in the design of RL algorithms, such as the Markov property.
In practice, these engineered action spaces could consist of multiple control loops and can abstract some physical phenomena.
There exist many options for action spaces that encapsulate robot control algorithms.

In this paper, our aim is to understand the role of the action space in manipulation learning.
\hl{We are interested in: 
\begin{enumerate}[label=(\roman*)]
    \item the exploration properties of different spaces,
    \item the emerging properties of policies trained with different low-level controllers,
    \item the gap created because of the choice of action space,
    \item the transferability of policies trained in different control spaces to the real world.
\end{enumerate}
}
We design a large-scale study to quantify these different aspects.
We include 13 different action spaces that we evaluate using multiple metrics.
measuring success rate transfer, the usability of resulting behaviors, and the gap introduced.
\section{Related Work}
\hl{\textbf{Action Spaces for Robot Learning.}} Recent work has shown that the choice of action space can play an important role in the success and performance of manipulation~\cite{martin2019variable, alles2022learning}, flying~\cite{kaufmann2022benchmark}, and locomotion policies~\cite{peng2017learning, duan2021learning}.
\hl{For manipulation, multiple action spaces for manipulation allow the policy to control its interaction with the environment, either by some direct means of force application~\cite{beltran2020learning,ulmer2021learning} or implicitly via impedance control~\cite{martin2019variable, luo2019reinforcement, bogdanovic_learning_2020}.}
Several methods explored the use of movement primitives in the action space~\cite{bahl2020neural,aljalbout2021learning}.
\hl{Another paradigm involves parameterized skills as discrete actions~\cite{sharma2020learning, chitnis2020efficient,dalal2021accelerating}.}
The majority of sim-to-real works use configuration space control~\cite{xie2020learning,hwangbo2019learning,akkaya2019solving,handa2023dextreme,tang2023industreal}.
The reason behind this is yet unclear.
Ganapathi et al.~\cite{ganapathi2022implicit} propose an approach to implicitly combine configuration and task-space actions using a differentiable forward kinematics model.
\hl{Recent approaches propose learning latent action spaces, aiming to reduce control dimensionality to the task manifold~\cite{zhou_plas_2020,allshire2021laser} or serve a different purpose such as coordinating multi-robot tasks~\cite{aljalbout2023clas}.}

\hl{\textbf{Action Space Studies.}}
There have been multiple studies concerning the action space in robotics~\cite{varin2019comparison, kaufmann2022benchmark, peng2017learning}.
These studies span robotic applications including locomotion~\cite{peng2017learning}, manipulation~\cite{varin2019comparison}, and flying robots~\cite{kaufmann2022benchmark}.
However, the majority of these studies focus on a very limited set of action spaces or tasks.
Multiple of them study this problem in simulation only~\cite{peng2017learning,varin2019comparison, martin2019variable}.
\hl{For instance, Mart{\'\i}n-Mart{\'\i}n et al.~\cite{martin2019variable} compare 4 action spaces for manipulation in simulation only, as part of the evaluation of their proposed action space.}
In contrast, Kaufmann et al.~\cite{kaufmann2022benchmark} study the transfer of flying policies from simulation to the real world with three different control spaces.
In this work, we study the role of the action space for learning manipulation policies in simulation, and also their transfer to the real world.
\hl{Our study includes 13 action spaces, spanning combinations of commonly used design characteristics in manipulation.}

\begin{figure}
    \centering
    \includegraphics[width=\linewidth]{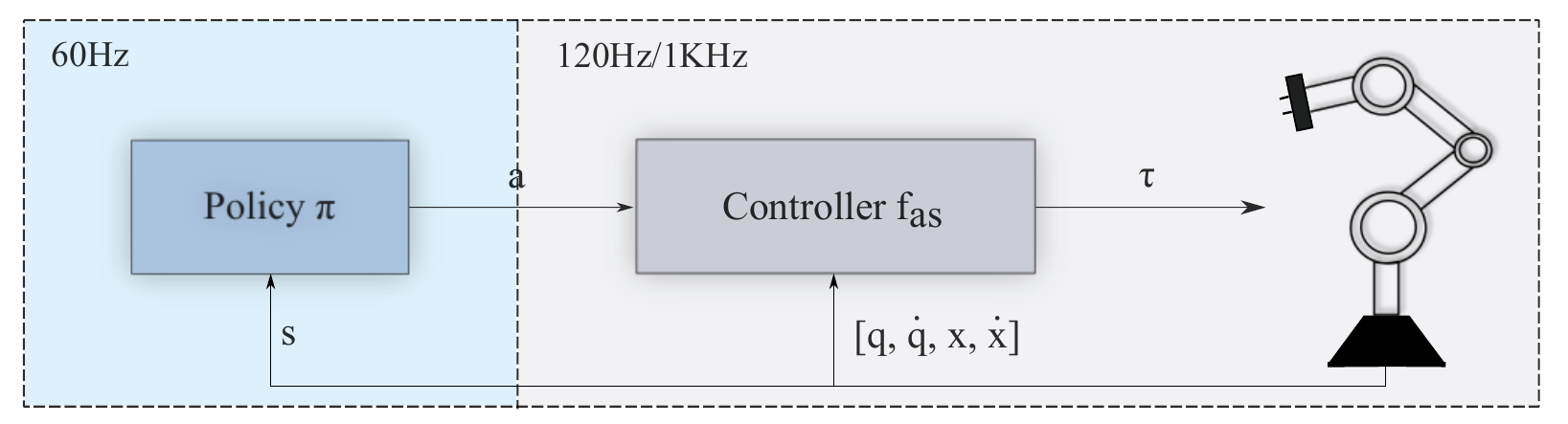}
    \caption{The policy outputs an action $a$, which is then transformed into joint torques $\tau$ using a select controller $f_{as}$. The policy and controller receive feedback from the environment. Each action space is defined by the choice of the controller and the way the action is treated in the controller. The policy runs at a 60\,Hz frequency and the controller runs at a frequency of 120\,Hz and 1\,kHz in the simulation and the real world, respectively. }
    \label{fig:actspace}
    \vspace{-0.5cm}
\end{figure}
\section{Methods}
In robot learning, the choice of action space involves the choice of a controller and the frequency and limits of the policy outputs.
The frequency directly defines the reactive capabilities of the policy, and the limits are typically given by the choice of controller and robot.
In this paper, we aim to study low-level control policies.
We fix the frequency of all action spaces to 60\,Hz.
We use the limits that are given by the robot for the corresponding control variables.

\subsection{Reinforcement Learning}

RL  automates policy learning by  maximizing the cumulative reward in a given environment~\cite{sutton2018reinforcement}.
Tasks are usually formulated as Markov Decision Processes~(MDP). A finite-horizon, discounted MDP is defined by the tuple $\mathcal{M} = (\mathcal{S}, \mathcal{A}, \mathcal{P}, r, \rho_0, \gamma, T )$, where $\mathcal{S}$ and $\mathcal{A}$ are the state and action spaces respectively, $\mathcal{P}: \mathcal{S} \times \mathcal{A} \to \mathcal{S}$ is the transition dynamics, and $r: \mathcal{S} \times \mathcal{A} \to \mathbb{R}$ the reward.
We have an initial state distribution $\rho_0$, a discount factor $\gamma \in [0,1]$, and a horizon length $T$. 
The optimal policy $\pi : \mathcal{S} \to P(\mathcal{A})$ maximizes the expected discounted reward:
\begin{equation}
\label{eq:RL}
J(\pi) =  E_{\pi} \left[ \sum_{t=0}^{T-1} \gamma^t r(s_t, a_t) \right].
\end{equation}
In this study, we use the proximal policy optimization~(PPO) algorithm~\cite{schulman2017proximal} for learning manipulation policies.
This choice is based on the popularity of this algorithm in recent works on sim-to-real transfer~\cite{tan2018sim,akkaya2019solving,chebotar2019closing,handa2023dextreme,tang2023industreal}.

\subsection{Action Spaces}
\label{eq:actionspace}

For arm manipulation, one of the most native RL action spaces is one that expects \textbf{joint torque~(JT)} commands.
This corresponds to $\tau=a$, where $\tau$ is the vector of joint torques.
Such an action space gives the policy full control over the robot.
When given the right observations, such a policy can internally learn to control the motion and forces exerted by the robot.
This action space was popular in early works using deep RL~\cite{wahlstrom2015pixels,lillicrap2015continuous}.
However, learning a policy with this action space can be very complicated since the policy would need to either understand or implicitly handle both the kinematics and dynamics of the robot to fulfill the task. 
This is due to the fact that the reward is typically expressed using task space properties.

Alternatively, a low-level controller $f_\mathrm{as}: \mathcal{A} \longmapsto T$ can be integrated in the action space to convert higher-level policy actions into the space of joint torques $T$ of the robot.
This concept is illustrated in Figure~\ref{fig:actspace}.

\textbf{Configuration action spaces} consist of all action spaces that expect a configuration-space action from the policy.
At the center of all these action spaces is the same controller, namely a joint impedance controller (JIC)\hl{~\cite{hogan1985impedance}}. 
JIC regulates the behavior of a robot manipulator's joints and allows specifying desired stiffness, damping, and inertia characteristics for each joint.
This allows the robot to be compliant with its environment.
The JIC control law is the following:
\begin{align}
    \label{eq:jic}
    \tau &= f_\mathrm{JIC}(a)\\
     &= K (q_d - q) + D (\dot{q_d} - \dot{q}),
\end{align}
where $\tau$ denotes the commanded joint torque, $q_d$ and $\dot{q_d}$ denote the desired joint positions and velocities respectively, and $q$ and $\dot{q}$ denote the actual joint positions and velocities of the robot given as feedback.
$K$ and $D$ are the stiffness and damping matrices.
\hl{Note that we omit the gravity vector from our controller equations for clarity.}
In practice, we use isotropic gains, 
i.e., these matrices are diagonal.
Given this control law, we can define two different base configuration action spaces:
\begin{itemize}[leftmargin=*, itemsep=0pt]
    \item \textbf{Joint Velocities~(JV)}: sets $f_\mathrm{JV}\leftarrow f_\mathrm{JIC}$, and $\dot{q_d}=s(a)$ and $q_d$ is \hl{computed as} a first-order integration of $\dot{q_d}$;
    \item \textbf{Joint Position~(JP)}: sets $f_\mathrm{JP}\leftarrow f_\mathrm{JIC}$, and $q_d=s(a)$ and $\dot{q_d}$ is \hl{computed as}  a first-order differentiation of $q_d$.
\end{itemize}
$s(a)$ is a function that scales the action vector to the limits of the corresponding output vector.

\textbf{Task action spaces} are defined using variables that are in the task/Cartesian space of the robot.
In the absence of accurate system identification, we can use the following Cartesian impedance controller \hl{as described in~\cite{khatib1987UnifiedApproachMotion}}:
\begin{equation}
    \tau = J(q)^T \bigl( K (x_d - x) + D(\dot{x_d}-\dot{x})\bigr),
\end{equation}
where $J(q)$ is the Jacobian matrix for the current robot configuration $q$, relating joint velocities to Cartesian velocities, $x_d$ and $\dot{x_d}$ are the desired Cartesian poses and velocities, $x$ and $\dot{x}$ are the current Cartesian poses and velocities respectively.
However, this formulation can be tricky to use when the $x_d$ and $\dot{x_d}$ are generated by an RL policy.
This is due to the non-smooth nature of RL action trajectories.
Of course, this problem can be handled by introducing interpolators or cubic spline fitting\hl{~\cite{martin2019variable,robosuite2020}}.
But such solutions involve multiple design choices and hyperparameters, meaning that an ideal solution is task-specific. 
Instead, we transform the Cartesian actions into joint velocities, and then use a joint impedance controller.
We found this approach to be very good at handling the non-smooth policy actions, without introducing any additional sim-to-real gap.
We have two base task action spaces:
\begin{itemize}[leftmargin=*, itemsep=0pt]
    \item \textbf{Cartesian Velocities~(CV)}: sets $\dot{x_d}\leftarrow s(a)$, transforms $\dot{x_d}$ into $\dot{q_d}$ using inverse kinematics (IK), and then uses $f_\mathrm{JIC}$ in the same fashion as in the JV action space;
    \item \textbf{Cartesian Position~(CP)}: sets $x_d\leftarrow s(a)$, and then uses a proportional control to obtain $\dot{x_d}$ from $x_d$.
    This step naturally results in smooth $\dot{x_d}$.
    Given $\dot{x_d}$, this action space proceeds as in the CV action space.
\end{itemize}
We use the pseudoinverse IK method with a null-space controller that pushes the joints toward their default positions.
\hl{We represent the orientation actions differently for these two spaces. In CP, we use the 6D representation presented in~\cite{zhou2019continuity}.
In CV, we use the Euler representation.}

\textbf{Delta action spaces} are based on the base action spaces defined previously.
In contrast to the base action spaces, delta action spaces set the control targets $v_d$ relative to the current system feedback or to the control targets of the previous policy step.
This distinction creates two classes of delta action spaces:
\begin{itemize}[leftmargin=*, itemsep=0pt]
    \item \textbf{One-step Integrator~(OI$\Delta$)}: uses the robot feedback $v$ to set $v_d\leftarrow v + c \cdot a \cdot dt$;
    \item \textbf{Multi-step Integrator~(MI$\Delta$)}: recurrently sets $v_d\leftarrow v_d + c \cdot a \cdot dt$,
\end{itemize}
\hl{where OI$\Delta$ is quite common~\cite{brohan2023RT2VisionLanguageActionModels,lee2020MakingSenseVision} and an action space similar to MI$\Delta$ was proposed in~\cite{tang2023industreal}.}
Depending on the choice of base action space, $v_d$ is a control target vector, and can correspond to $q_d$, $\dot{q_d}$, $x_d$, or $\dot{x_d}$.
Similarly, $v$ is a control feedback vector, and can correspond to $q$, $\dot{q}$, $x$, or $\dot{x}$.
$dt$ is the step duration and $c$ is a positive constant hyperparameter.
Instead of the scaling performed in non-delta base action spaces, we clip the target $v_d$ to the limits of the corresponding output space after updating it.
This means that each base (configuration or task) action space has two additional variants, resulting in 12 action spaces, not including the joint torques action space.
Despite relying on the base action spaces, the delta variants have their unique properties.
For instance, due to the relative changes in the control targets, the magnitude of the target change in one step is bound by $c \cdot dt$ given that $a \in [-1,1]$.
This property can be helpful in imposing smoothness constraints on control target trajectories, even if policy output is unconstrained.
If we set $c$ to be the positive bound of the derivative of the corresponding control feedback variable $v$, each delta action space would approximately be equivalent to an action space of a higher-order derivative than its base action space.
For instance, a delta joint velocity action space would be approximately similar to a joint acceleration action space.
However, when sampling actions $a$ from a uniform distribution, the resulting distribution of control targets can differ between a delta action space and the base action space it approximates (e.g.\ $\Delta$JP and JV).
The clipping can also lead to more probability mass on the borders of the $v_d$ space.

\subsection{Metrics}
\label{seq:metrics}
For each action space, we aim to assess its training performance, resulting sample efficiency, emerging properties such as the usability in the real-world environment, and the sim-to-real gap it creates.
Therefore, we propose multiple metrics to quantify these different aspects.
For assessing training performance, we look at the \textbf{episodic rewards~(ER)} in simulation.
This metric can also show us the sample efficiency of the different action spaces.

To assess the emerging properties of each policy, we look at the number of times it violates robot constraints, such as the joint acceleration and jerk constraints.
This is especially important since simulated environments rarely have any mechanisms for enforcing these constraints.
In contrast, real-world robot control implementations can include rate limiters, which ensure that the control targets would not result in violations of these constraints.
A policy trained in simulation without such mechanisms could learn to violate them for the sake of exploration.
Implementing these mechanisms in simulation, on the other hand, can also hinder the policy training since these mechanisms typically break the Markov assumption.
To quantify this property, we use the \textbf{expected constraints violations~(ECV)}  defined as
\begin{equation}
    \mathrm{ECV}(\pi) = E_{s,a \sim \pi} \left[ \mathds{1}\left(\sum_{c \in \mathcal{C}} c(s,a) > 0 \right) \right],
    \label{eq:ECV}
\end{equation}
where $\mathds{1}$ is an indicator function, $\mathcal{C}$ is the set of all constraints, and each constraint function $c(s,a)$ returns 1 for violated constrained and 0 otherwise. 
Furthermore, we report the \textbf{normalized tracking error~(NTE)} as a measure of the feasibility of the policy's actions in the environment,
\begin{equation}
    \mathrm{NTE}(\pi) = E_{s,a \sim \pi} \frac{|v_{d,t} - v_{t+1} |}{v_\mathrm{max} - v_\mathrm{min}} ,
    \label{eq:NTE }
\end{equation}
where $v_\mathrm{min}$ and $v_\mathrm{max}$ are respectively the lower and upper bounds of the control variable $v$.
NTE is useful for analysing why certain action spaces result in better transfer.
A high value means that the policy outputs actions that are hard to achieve in one control step.
This could create an additional sim-to-real gap since control targets that are not fulfilled in one step can be tracked differently in simulated and real environments due to the gap in dynamics.
We also report the \textbf{task accuracy~(ACC)} measured by the Euclidean distance to the goal.
This metric gives a more detailed view of the performance of a policy than just the success rate.

To assess the sim-to-real gap of each action space we report the \textbf{offline trajectory error~(OTE)} in configuration and task spaces.
This metric measures the joint trajectory error when replaying, in simulation, the actions produced by the policy when queried in the real world,
\begin{equation}
    \mathrm{OTE}(\pi) = E_{a, q_\mathrm{real} \sim \mathcal{D}_\mathrm{real}} |q_\mathrm{sim} -q_\mathrm{real} | ,
    \label{eq:OTE}
\end{equation}
where $a$ and $q_\mathrm{real}$ are actions and joint configurations that are sampled from the dataset $\mathcal{D}_\mathrm{real}$. 
The latter is collected by playing a policy $\pi$ in the real-world, and $q_\mathrm{sim}$ is the joint configuration obtained when executing $a$ in simulation in an open-loop fashion, i.e.\ without a policy.

\section{Experiments}

\begin{figure}
    \centering
    \includegraphics[width=0.49\linewidth]{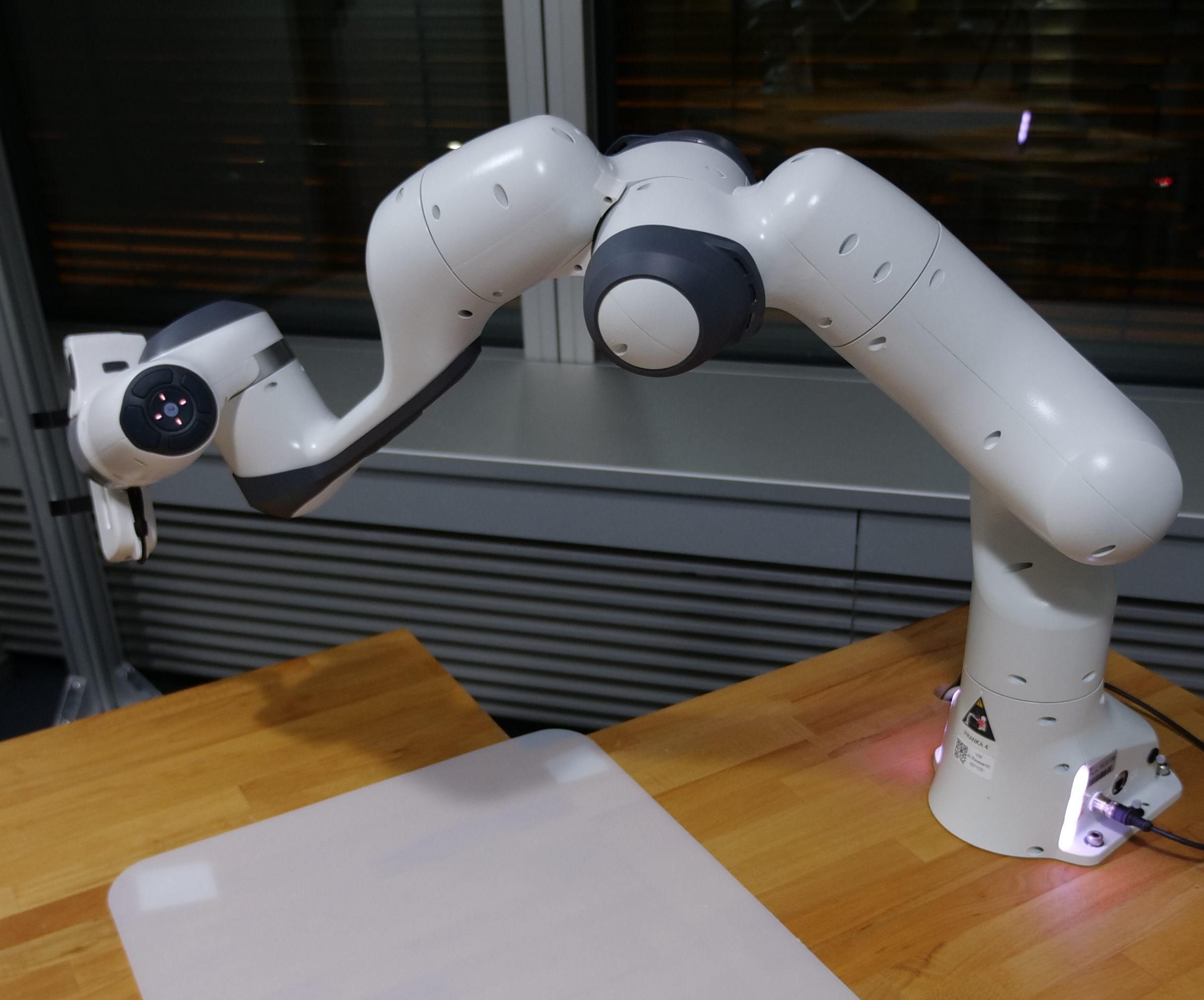}
    \hfill
    \includegraphics[width=0.49\linewidth]{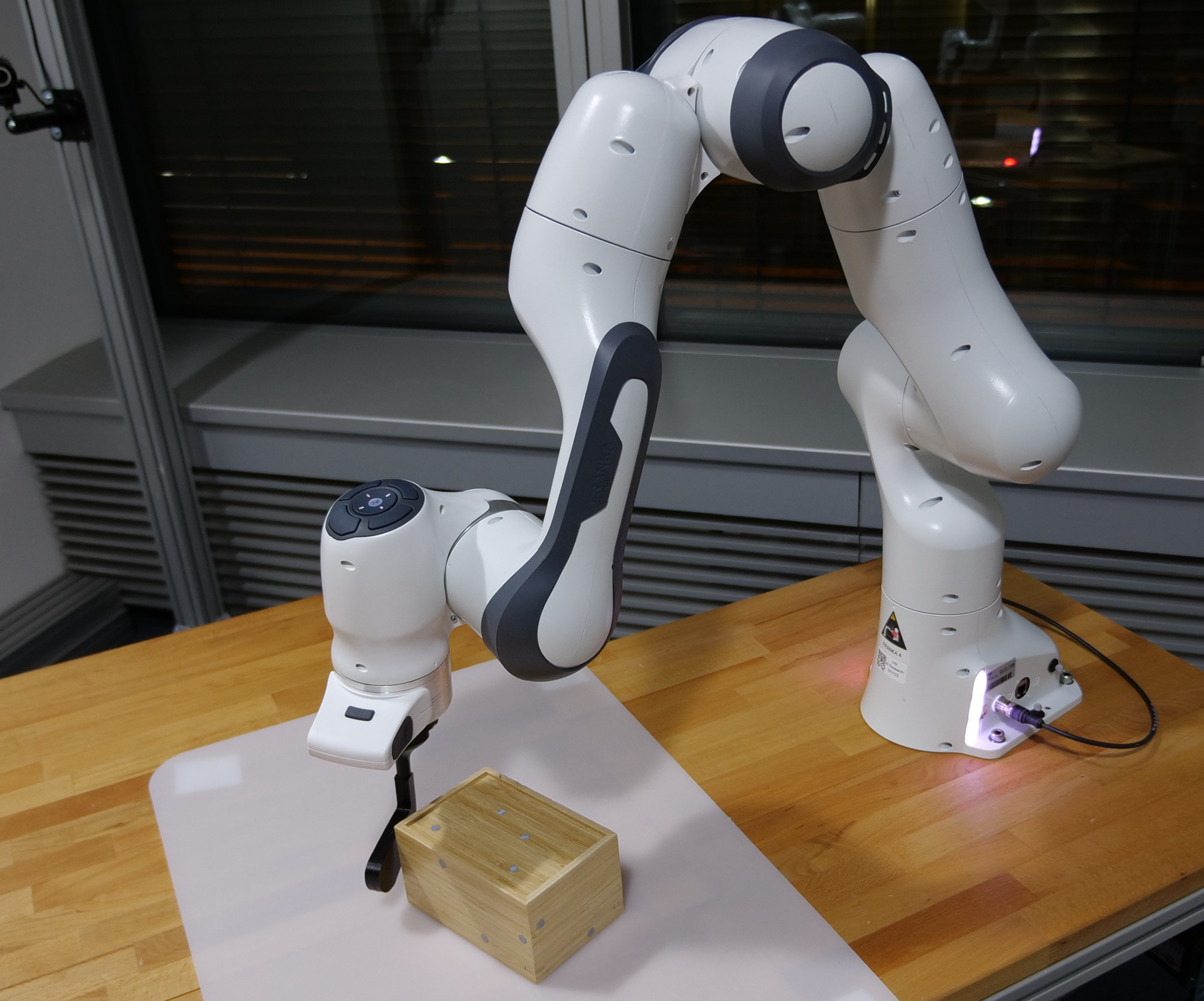}
    \caption{We show our real-world robot setup for the reaching (left) and the pushing (right) tasks.}
    \label{fig:robot-setup}
    \vspace{-0.6cm}
\end{figure}

We design experiments to understand the role of different characteristics of the action space on learning efficiency, sim-to-real gap, emerging properties, and sim-to-real transfer.
Throughout this section, we pose different questions related to these topics, and answer them to the best of our capability based on empirical analysis.
\subsection{Experimental Setup}

To answer these questions we evaluate all action spaces on two arm manipulation tasks, using the 7-degree-of-freedom Franka Emika Panda robot.
The first task is goal reaching.
At the beginning of each episode, a Cartesian goal is sampled in the workspace of the robot, and the policy needs to move the end-effector towards that goal.
This task is ideal for studying the behavior of policies from all action spaces in the lack of force interactions with the robot's environment.
The second task is object pushing.
At the beginning of each episode, a goal position is sampled in a predefined area on the table. 
The policy needs to push a wooden box towards that goal.
This task involves moving an external object.
Unlike the reaching task, pushing requires physical interaction with the environment.
It allows us to understand the reactive capabilities of each action space and whether it creates any sim-to-real gap that hinders the transfer of the interaction behavior.
During policy training for pushing in simulation, we perform domain randomization on the box's friction and mass parameters.
\hl{For each task, we train 5 different policies for each action space variant, using random seeds. 
We use a fixed number of training epochs for all policies and we evaluate the best 3 policies on the real-world setup.}
The real-world setup for both tasks is shown in Figure~\ref{fig:robot-setup}.
In both tasks, the observation space of the policy consists of joint positions, joint velocities, end-effector Cartesian position, and the goal position.
Additionally, in the pushing task the policy has access to the object's position and orientation.
We train PPO policies in a simulated environment, using NVIDIA's Isaac-sim simulator \hll{and the PhysX engine. We} \hl{leverage the robot model developed for the cabinet opening task} in~\cite{makoviychuk2021isaac}, \hll{and scan the woodbox object used in the real-world environment to use the meshes in our simulation.} \hl{Self-collisions are not enabled during training.}

\begin{figure*}
    \centering
    \includegraphics[width=\linewidth]{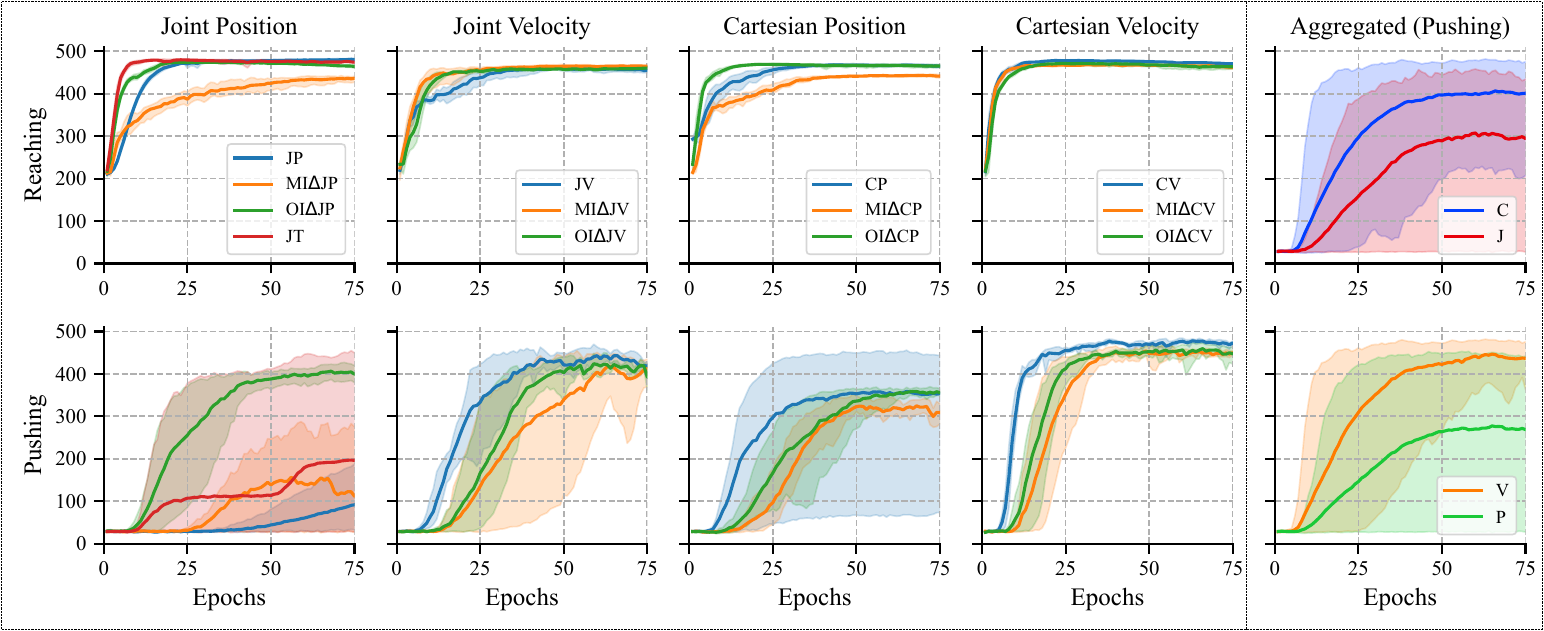}
    \caption{
    \hl{    
    We show the episodic reward (ER) obtained during training in simulation for all the studied action spaces.
    The first four columns show the learning curves grouped by action spaces, whereas the last column shows aggregated results comparing joint (J) and Cartesian (C) action spaces as well as position based (P) and velocity based (V) action spaces for the pushing task.
    Delta ($\Delta$) action spaces are labeled as either one-step (OI) or multi-step (MI) target integration.
    We added joint torque (JT) control to the first column.
    The shaded region represents the range between the 5th and 95th percentiles.
    }
    }
    \vspace{-0.2cm}

\label{fig:sim_reward}
\end{figure*}

The results discussed in this paper are based on 250 agents with different action spaces and different tasks.
However, prior to the final training runs (of the 250 agents), we optimized hyperparameters, such as learning rates or different reward scales and terms, to guarantee consistency and a fair comparison over all action spaces.
We matched the simulation to the real environment to the best of our knowledge, eliminating sim-to-real gaps where possible.
This includes, for example, matching the stiffness of the impedance controllers, sharing implementations for each action space across both environments, and finding good velocity limits that allow for safe transfer of policies.
This tuning process required training more than 20,000 agents in simulation and testing more than a 1500 agents in the real world.
After training, we evaluate the learned policies in both tasks in simulation and in the real world.
For reaching, we use a fixed grid of target goals that span the feasible workspace of the robot.
For pushing, we randomly sample goal positions and use the object position from the previous episode as the initial one from the new run.
We only manually reset the object's position whenever it gets outside of the predefined area it was trained to operate in or when the policy fails to push the object away from its starting location once.
We introduced the latter intervention to avoid heavily skewing the data by randomly sampling a difficult-to-handle initial object position.
In both training and evaluation, we run the simulation at 120\,Hz and we use action repeat to work with the policy output at 60\,Hz.
\hl{We originally experimented with higher frequency joint torques policies, but found for our setup no benefit in going beyond 60\,Hz.}
On the real robot, we need an additional low-level safety controller running at 1\,kHz, which also repeats the actions from the policy.
In all action spaces except torque we additionally use a 5\,Hz low-pass filter and a rate limiter to conform to motion limits given by the robot.
We tried implementing the rate limiter in simulation as well, however, the policy learning deteriorated massively, which we attribute to the rate limiter breaking the Markov property.
\hl{For the reaching policy, we use a 4-layer feed-forward neural network with (512, 256, 128, 64) units for all action spaces. For pushing, we use 5-layer networks with (1024, 1024, 512, 256, 128) units. Furthermore, for both tasks, torque policies use 2 additional layers with 1024 neurons.}

\subsection{Results}
\hl{
We study different properties of action space selection in robotic manipulation and sim-to-real transfer.
}
\\
\\
\textbf{Does the choice of action space affect the exploration behavior during training in simulation?}
\\
We first examine the training performance in simulation. 
The results can be seen in the Figure~\ref{fig:sim_reward}.
The episodic rewards are not directly representative of the success rate of the policies.
This can be seen when comparing the rewards from Figure~\ref{fig:sim_reward} to the success rates shown in Table~\ref{tab:sim2real}.
Therefore, we mostly focus on sample efficiency in the current analysis.
Since different action spaces have different characteristics, we aim to understand the global effect of these characteristics on the sample efficiency during training.
First, we compare Cartesian and joint action spaces. 
In the reaching task, both action space groups behave almost identically.
However, in the pushing task, Cartesian action spaces seem to have an advantage in terms of sample efficiency and maximum reached reward, as can be seen in the top right plot in Figure~\ref{fig:sim_reward}.
This result is most likely due to the spatial nature of the pushing task, which gives Cartesian action spaces a natural advantage in exploration.

We next compare the order of derivative of the action space, i.e.\ positions vs.\ velocities vs.\ torques.
Among joint action spaces, joint velocity and its derivatives have the overall best performance in both tasks. 
These action spaces show better sample efficiency and converge to higher reward regions than their counterparts in the joint action space group as can be seen in the bottom right plot in Figure~\ref{fig:sim_reward}.
JP reaches the highest reward in reaching but struggles massively in pushing.
The same tendency can be observed for Cartesian action spaces, i.e.\ velocity action spaces perform better in terms of sample efficiency and final rewards.
The joint torque action space is the fastest one to converge in the reaching task, however, it fails to solve the pushing task reliably within the same data budget. \hl{Training pushing policies in this action space for a longer time does converge, but is very expensive.}

We also compare base action spaces with the two different kinds of delta action spaces.
We observe that multi-step integration delta action spaces consistently perform the worst, while one-step methods seem to have a slight advantage.
This tendency is less pronounced in the velocity action spaces, with MI$\Delta$JV being consistently one of the best-performing joint action spaces in both tasks in simulation.
Hence, the current simulation data does not conclusively favor any of these characteristics (non-delta, OI and MI).
Overall, we note that Cartesian velocity~(CV) is the best-performing action space in simulation.
The worst one is multi-step-integration joint position action space.
\begin{table*}[]
    \centering
    \caption{Sim-to-real transfer evaluation. We evaluate the success rate~(SR) in simulation and the real-world environment, the task accuracy~(ACC), the expected constraints violations~(ECV), and the offline trajectory error~(OTE).}
    \label{tab:sim2real}
    \sisetup{detect-weight=true}
    \begin{tabular}{
        l
        S[table-format = 3.0]@{\,\,\,\( \pm \)\,}
        S[table-format = 1.0]
        S[table-format = 3.0]@{\,\,\( \pm \)\,}
        S[table-format = 1.0]
        S[table-format = 1.2]@{\,\( \pm \)\,}
        S[table-format = 1.2]
        S[table-format = 3.0]@{\,\( \pm \)\,}
        S[table-format = 2.0]
        S[table-format = 1.2]@{\,\( \pm \)\,}
        S[table-format = 1.2]
        c
        S[table-format = 3.0]@{\,\,\( \pm \)\,}
        S[table-format = 1.0]
        S[table-format = 2.0]@{\,\,\,\( \pm \)\,}
        S[table-format = 1.0]
        S[table-format = 1.2]@{\,\( \pm \)\,}
        S[table-format = 1.2]
        S[table-format = 3.0]@{\,\( \pm \)\,}
        S[table-format = 2.0]
    }
\toprule
\multicolumn{1}{c}{\bfseries Action} & \multicolumn{10}{c}{\bfseries Reaching} & & \multicolumn{8}{c}{\bfseries Pushing}\\
\multicolumn{1}{c}{\textbf{space}} &
\multicolumn{2}{c}{\textbf{SR} (sim) $\uparrow$} &
\multicolumn{2}{c}{\textbf{SR} (real) $\uparrow$} &
\multicolumn{2}{c}{\textbf{ACC} [cm] $\downarrow$} &
\multicolumn{2}{c}{\textbf{ECV} $\downarrow$} &
\multicolumn{2}{c}{\textbf{OTE} [rad] $\downarrow$} & &
\multicolumn{2}{c}{\textbf{SR} (sim) $\uparrow$} &
\multicolumn{2}{c}{\textbf{SR} (real) $\uparrow$} &
\multicolumn{2}{c}{\textbf{ACC} [cm] $\downarrow$} &
\multicolumn{2}{c}{\textbf{ECV} $\downarrow$} \\
\midrule
JP              &   \bfseries 100 & \bfseries 0 &               6 &           6 &             1.76 &           0.14 &           100 &           0 &               0.22 &           0.01 & &                87 &          12 &              4 &           0 &              4.23 &           0.66 &           100 &           0 \\
OI$\Delta$JP    &   \bfseries 100 & \bfseries 0 &              86 &           9 &             1.59 &           0.16 &            14 &           1 &               0.45 &           0.01 & &                99 &           3 &             48 &          12 &              3.16 &           0.46 &            98 &           2 \\
MI$\Delta$JP    &   \bfseries 100 & \bfseries 0 &              69 &           2 &             2.14 &           0.13 &   \bfseries 0 & \bfseries 0 &     \bfseries 0.03 & \bfseries 0.00 & &                90 &           7 &             36 &           3 &              3.52 &           0.36 &   \bfseries 3 & \bfseries 1 \\
\midrule
JV              &   \bfseries 100 & \bfseries 0 &   \bfseries 100 & \bfseries 0 &   \bfseries 1.17 & \bfseries 0.11 &   \bfseries 0 & \bfseries 0 &     \bfseries 0.03 & \bfseries 0.00 & &                97 &           5 &             90 &           5 &              2.04 &           0.32 &            66 &           9 \\
OI$\Delta$JV    &   \bfseries 100 & \bfseries 0 &              97 &           5 &             1.35 &           0.13 &            13 &          11 &               0.18 &           0.01 & &                99 &           3 &             88 &          13 &    \bfseries 1.56 & \bfseries 0.13 &            75 &           4 \\
MI$\Delta$JV    &   \bfseries 100 & \bfseries 0 &              93 &           9 &             1.98 &           0.08 &            37 &          37 &     \bfseries 0.03 & \bfseries 0.00 & &     \bfseries 100 & \bfseries 0 &   \bfseries 96 & \bfseries 4 &              1.66 &           0.11 &            55 &          25 \\
\midrule
JT              &   \bfseries 100 & \bfseries 0 &              64 &          27 &             1.97 &           0.25 &            22 &           6 &               0.52 &           0.03 & &                40 &          49 &       \text{-} &    \text{-} &             \text{-} &    \text{-} &            89 &          21 \\
\midrule
CP              &              99 &           2 &              25 &           4 &             1.71 &           0.29 &            26 &           9 &               0.16 &           0.01 & &     \bfseries 100 & \bfseries 0 &             57 &           9 &              2.34 &           0.39 &            92 &           3 \\
OI$\Delta$CP    &              89 &           2 &              44 &          17 &             2.02 &           0.32 &            21 &           1 &               0.27 &           0.01 & &                99 &           3 &             55 &          17 &              2.86 &           0.49 &            35 &          16 \\
MI$\Delta$CP    &   \bfseries 100 & \bfseries 0 &              68 &           2 &             1.85 &           0.08 &            13 &           4 &               0.28 &           0.01 & &     \bfseries 100 & \bfseries 0 &             73 &           7 &              2.36 &           0.14 &            99 &           0 \\
\midrule
CV              &   \bfseries 100 & \bfseries 0 &              43 &          13 &             1.76 &           0.41 &            12 &           2 &               0.35 &           0.02 & &                99 &           3 &             73 &           9 &              2.69 &           0.58 &            93 &          10 \\
OI$\Delta$CV    &              97 &           2 &              24 &           2 &             2.13 &           0.13 &            17 &           3 &               0.40 &           0.01 & &                99 &           3 &             61 &          19 &              2.67 &           0.30 &            99 &           0 \\
MI$\Delta$CV    &              99 &           2 &              58 &           8 &             1.82 &           0.08 &             4 &           0 &               0.13 &           0.01 & &                86 &           5 &             74 &           4 &              2.84 &           0.45 &             6 &           2 \\
\bottomrule
\end{tabular}
    \vspace{-0.3cm}
\end{table*}
\\
\\
\textbf{What properties naturally emerge due to the choice of action space?}
\\
After training the policies in simulation we evaluate them on a real robotic setup.
For the pushing task, we were not able to obtain joint torque policies that are safe to run in the real world without damaging the robot.
We made multiple attempts to produce safely deployable joint torque policies, for instance, by introducing different penalties or increasing the policy's control frequency.
Despite all efforts, all joint torque policies were very jerky or aggressive when deployed on the real robot.
The main difficulty was to obtain policies that are safe to deploy in a task where the end effector needs to remain in close proximity to a rigid surface, e.g.\ the table.
Therefore, we exclude the joint toque action space from our real-world pushing experiments.
For all other action spaces, we look at their sim-to-real transfer capabilities.
The reaching task was less safety-critical. 
Hence, we managed to evaluate torque policies on real-world reaching.
In Table~\ref{tab:sim2real}, we report the metrics introduced in Section~\ref{seq:metrics} to quantify the sim-to-real gap and the performance in the real world.
One common challenge when learning manipulation skills is to obtain smooth policies that do not violate the velocity, acceleration, and jerk constraints of the robot.
Based on the results in Table~\ref{tab:sim2real}, we observe that different action spaces yield different ECV metrics.
JV and its derivatives (OI$\Delta$JV and MI$\Delta$JV) have on average the lowest ECV score in both tasks.
Nominal JP results in the highest possible ECV. 
This means that deploying this action space in the real world would always result in some form of constraint violation.
In the absence of safety mechanisms (such as rate limiters and low-pass filters) these violations can be very harmful. 
If such mechanisms are implemented, this behavior would increase the sim-to-real gap further because of how these violations trigger the safety mechanisms.
In contrast, MI$\Delta$JP, seems to have the lowest ECV, which was also evident when evaluating this action space in the real world.
\begin{figure}
    \centering
    \includegraphics[width=0.95\linewidth]{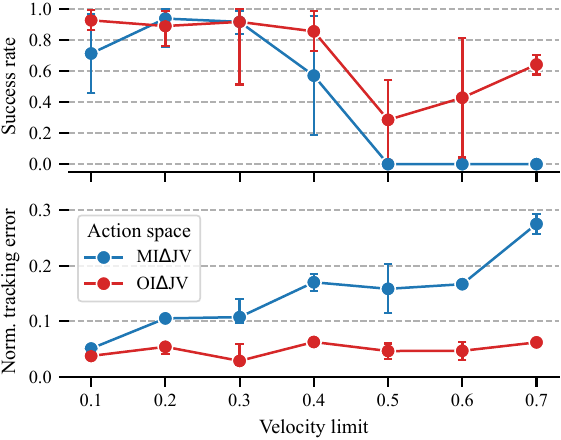}
    \caption{%
        Robustness of delta action spaces to the velocity limit hyperparameter in the pushing task on our real setup.
        We compare one-step and multi-step integration.
        \hl{The error bars indicate the 5th and 95th percentiles.}}
    \label{fig:OIMIrobustness}
    \vspace{-0.62cm}
\end{figure}
\\
\\
\textbf{Does the action space affect the sim-to-real gap?}
\\
First, we look at the offline trajectory error in the reaching task, which gives us a proxy measure of the sim-to-real gap.
We observe that the OTE varies tremendously from one action space to another, despite the fact that the data used to compute this measure are based on the same starting and goal positions in all action spaces.
This confirms that \textit{the choice of action space does indeed contribute to the sim-to-real gap}.
Looking closer at the results, we observe that JT exhibits the highest OTE. 
This is due to the behavior of this action space being dictated solely by the dynamics of the robot, which is different in simulation and in the real world.
Unlike all the other action spaces, JT does not include any feedback loops outside of the policy.
Based on this result, one would expect that more feedback loops should help reduce the sim-to-real gap.
However, the opposite can be seen in the data.
For instance, Cartesian action spaces have on average one additional feedback loop compared to the joint action spaces. 
Their OTE is, however, higher on average.
This is due to the fact that feedback loops have different effects in simulation than in the real world.
In turn, this means that a good, highly-reactive feedback loop is beneficial to overcome the dynamics gap, but adding more could potentially contribute further to the sim-to-real gap.
Comparing OI$\Delta$ and MI$\Delta$ action spaces we notice that the latter consistently have a smaller OTE.
Their OTE is even smaller than the corresponding base action spaces.
This effect is potentially due to the integral term embedded in these action spaces.
Executing the same actions results in the same final goal given to the lower-level feedback loops, which is unique compared to all other action spaces.
\\
\\
\textbf{Which action space characteristics are good for sim-to-real transfer? }
\\
We compare the success rates in simulation and the real world.
We observe that \textit{the success rate in simulation is not directly reflected in the real world}.
This is clear when comparing the ordering of success rates in both domains.
Furthermore, we notice that certain action space characteristics are clearly advantageous for sim-to-real transfer.
For instance, velocity-based action spaces tend to keep a high success rate when transferred to the real world.
In contrast, position-based action spaces do not typically transfer well.
This is especially the case for JP, which loses almost all of its performance when transferred.
These results are consistent across both tasks.
Velocity-based action spaces are on average more accurate in fulfilling the task as can be seen when comparing the ACC score in Table~\ref{tab:sim2real}.

\begin{figure}
    \centering
    \includegraphics[width=0.97\linewidth]{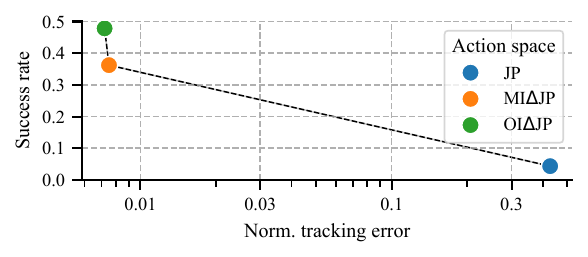}
    \vspace{-0.3cm}
    \caption{We compare joint position-based action spaces in the pushing task and show their influence on the normalized tracking error and the resulting effect on success rate in the real robot setup.}
    \label{fig:push-jp-compare}
    \vspace{-0.6cm}
\end{figure}

\begin{figure}
    \centering
    \includegraphics[width=0.95\linewidth]{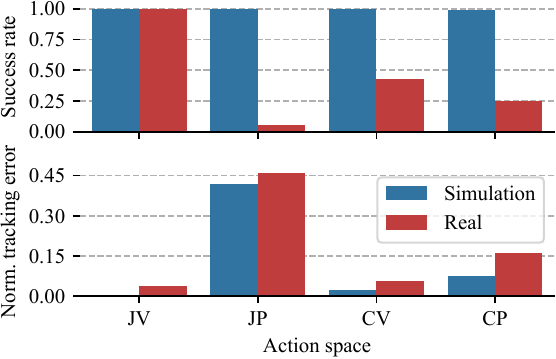}
    \vspace{-0.1cm}
    \caption{We show the influence of the normalized tracking error on sim-to-real transfer in terms of success rate for non-delta action spaces in the reaching task.}
    \vspace{-0.6cm}
    \label{fig:ntevanilla}
\end{figure}

Additionally, we can see that delta action spaces transfer better than their corresponding base spaces.
The difference between OI and MI delta action spaces depends on the task and the variants of the action space.
However, OI$\Delta$ action spaces required less tuning than MI$\Delta$ spaces and have shown to be more robust to the choice of hyperparameters.
This can be seen in Figure~\ref{fig:OIMIrobustness}.
When varying the velocity limits for $\Delta$JV action spaces, OI$\Delta$JV showed to be less sensitive to the chosen value.
We attribute this to the lower tracking error of this action space as shown in the bottom plot in Figure~\ref{fig:OIMIrobustness}.
OI$\Delta$ action spaces naturally lead to a lower tracking error since they integrate the policy actions into control targets based on the current feedback of the system.
In contrast, MI$\Delta$ action spaces integrate the policy actions into control targets based on the previous control target.
This in turn leads to a bigger gap between the control target and the state of the system, and hence a larger tracking error.
While a lower NTE helps make JV action spaces more robust to hyperparameters, it has an even larger effect on the best possible transfer performance in JP action spaces, as shown in Figure~\ref{fig:push-jp-compare}.
This tendency is also evident when comparing the base spaces, as shown in Figure~\ref{fig:ntevanilla}.
Finally, joint velocity show better transfer than Cartesian velocity action spaces, while the opposite is true for position-based action spaces.

In summary, our data shows that two characteristics mostly influence the transfer capability of an action space.
The first one is the order of the derivative of the control variables.
\textit{An action space that controls a higher order derivative transfers better.}
\hl{A clear exception to this finding is the torque action space, which suffers from the highest sim-to-real gap (as shown by the OTE in Table~\ref{tab:sim2real}) due to its direct reliance on the mismatched environment dynamics and lack of external feedback loops in comparison to the other action spaces.}
The second characteristic is the emerging tracking error of the action space's control variable(s).
Our data strongly indicates that \textit{an action space which yields or naturally limits the tracking error transfers better.}
This last property can be enforced by the action space design, for instance, by reducing the magnitude of jumps of the corresponding control targets.
The latter can be controlled by the scaling of the actions in delta action space, or by increasing the stiffness if the task allows.
For action spaces that do not naturally allow for controlling this property, one could attempt to enforce a smaller NTE by means of rewards/penalties on the actions' magnitude and smoothness.
However, based on our experience, the tuning process for the resulting additional hyperparameters (of such reward terms) can be very difficult.
\\
\\
\textbf{Is there a consistently best-performing action space? }
\\
Overall, we notice that \textit{joint velocity action spaces seem to have the best performance in sim-to-real transfer}.
These action spaces have on average the lowest OTE, ACC and ECV and transfer the best to the real world.
They also required the least tuning to work, making them the most suited action spaces for manipulation learning (among the studied options).
This result is consistent with our previous findings concerning favorable characteristics, i.e.\ JV-based action space benefits from a higher-quality feedback loop due to their configuration-space control, and can more easily generate high forces for interaction than position-based action space due to controlling a higher-order derivative.

\section{Conclusion}
We studied the role of the action space in learning robotic arm manipulation skills.
We designed a study that includes 13 action spaces that we used for training RL policies in simulation.
We looked at the exploration behavior under the different action spaces and observed that Cartesian spaces as well as ones that control a higher-order derivative have an exploration advantage. 
Furthermore, we observe that different action spaces lead to different emergent properties such as the safety of their transfer and their rate of constraints violations.
However, we notice that the success rate of policies in simulation does not dictate their performance in the real world.
Our study further shows that the characteristics that mostly affect the sim-to-real transfer are the order of derivative and the emerging tracking error of the action space.
Joint velocity-based action spaces show very favorable properties, and overall the best transfer performance.
Our results show that the choice of action space plays a central role in learning manipulation policies and the transfer of such policies to the real world.
\hl{
To validate the extent of our findings and recommendations to different tasks, observation spaces, robots, and hardware setups, more studies of this sort are required in the future.
}

\section*{ACKNOWLEDGMENT}
The authors would like to thank Alexandros Paraschos and Philip Becker-Ehmck for their help and support.

\balance
\bibliographystyle{ieeetr}
\bibliography{biblio}

\clearpage
\nobalance
\appendix

\section{Implementation Details}
We provide more details concerning our environments.
For the reaching task, we define the reward function
\newcommand{\norm}[1]{\left\lVert#1\right\rVert_2}
\begin{align*}
    r_{reach}(s_t, a_t) &= rew_{reach}(s_t, a_t) - pen_{reach}(s_t, a_t) \\
    rew_{reach} &= r_{dist} + r_{exact} \\
    pen_{reach} &= r_{vel}  +  r_{smooth} + r_{neutral} + r_{limit}\\
    r_{dist}(s_t, a_t) &=  \lambda_r \cdot \frac{1}{1 + \norm{s_t - g}^2} \\
    r_{exact}(s_t, a_t) &= \mathds{1}(\norm{s_t - g} < \epsilon)(\lambda_\epsilon + \frac{1}{1+100\dot{q}^2})\\
    r_{vel}(s_t,a_t) &= \lambda_q \cdot \norm{\dot{q}_{t}}^2 \\
    r_{neutral}(s_t, a_t) &= \lambda_n  \cdot \norm{q_{def}-q}\\
    r_{limit}(s_t, a_t) &= \lambda_l \cdot e^{-30(q-q_{lim})^2}\\
    r_{smooth}(s_t, a_t) &= \lambda_s  \cdot \norm{a_t - a_{t-1}} ,
\label{eq:reach_reward}
\end{align*}
where we use the Euclidean norm as a distance metric.
$g$ is the end-effector goal position, $q_{def}$ is the default joint positions vector, $\epsilon$ is a small positive constant, $\lambda_r$ and $\lambda_\epsilon$ scale the reach and exact reach reward, $\lambda_q$ and $\lambda_s$ are positive scalars for the penalties on the velocity magnitude and smoothness of the action respectively and $\mathds{1}$ is an indicator function, $lamda_n$ is a scalar for the penalties on divergence from the default joint position, and $\lambda_l$ is a scalar for the joint limit avoidance penalty. 
For pushing we have a different reward,
\begin{align*}
    r_{push}(s_t, a_t) &= rew_{push}(s_t, a_t) - pen_{push}(s_t, a_t) \\
    rew_{push} &= r_{dist} + r_{exact} + r_{push}  \\
    pen_{push} &= r_{vel}  +  r_{smooth} + r_{neutral} + r_{limit} + r_{col} \\
    r_{col}(s_t, a_t) &= \lambda_c \cdot \mathds{1}(z_{ee}<0.02),
\end{align*}
where $\lambda_c$ is a scalar for the table collision penalty, $z_{ee}$ is the end-effector's z-position, $r_{dist}$ and $r_{exact}$ use the object position as goal, and $r_{push}$ is defined exactly as $r_{dist}$, but measures the distance between the object's position and the pushing goal position.

\end{document}